\documentclass[letterpaper, 10 pt, journal, twoside]{IEEEtran}

\IEEEoverridecommandlockouts                              %

\usepackage{graphicx}
\usepackage{amsmath}
\usepackage{amssymb}
\usepackage{booktabs}
\usepackage{multirow}
\usepackage{xcolor}
\usepackage{colortbl}
\usepackage{subcaption}
\definecolor{LightCyan}{rgb}{0.8,1,1}
\usepackage{pifont}%
\newcommand{\cmark}{\ding{51}}%

\usepackage[capitalize]{cleveref}
\crefname{section}{Sec.}{Secs.}
\Crefname{section}{Section}{Sections}
\Crefname{table}{Table}{Tables}
\crefname{table}{Tab.}{Tabs.}

\usepackage{xspace}
\newcommand*{\eg}{e.g.\@\xspace}
\newcommand*{\ie}{i.e.\@\xspace}
\newcommand*{\etal}{et al.\@\xspace}

\usepackage{lipsum}  

\makeatletter
\newcommand*{\etc}{%
    \@ifnextchar{.}%
        {etc}%
        {etc.\@\xspace}%
}
\makeatother

\newcommand*{\precision}{\textcolor{orange}{P}\@\xspace}
\newcommand*{\recall}{\textcolor{cyan}{R}\@\xspace}
\newcommand*{\fscore}{\textcolor{purple}{F}\@\xspace}

\newcommand{\myparagraph}[1]{\vspace{2pt}\noindent{\bf #1}}

\title{
SupeRGB-D: Zero-shot Instance Segmentation \\in Cluttered Indoor Environments
}

\author{Evin P{\i}nar \"{O}rnek$^{1}$, Aravindhan K Krishnan$^{2}$, Shreekant Gayaka$^{2}$, \\ Cheng-Hao Kuo$^{2}$, Arnie Sen$^{2}$, Nassir Navab$^{1}$, Federico Tombari$^{1,3}$%
\thanks{Manuscript received: December, 23, 2022; Revised March, 15, 2023; Accepted April, 13, 2023.}
\thanks{This paper was recommended for publication by Editor Cesar Cadena Lerma upon evaluation of the Associate Editor and Reviewers' comments.}
\thanks{Work partially done during an internship at Amazon.}
\thanks{$^{1}$Technical University of Munich, Germany
        {\tt\small \{evin.oernek, nassir.navab, federico.tombari\}@tum.de} }%
\thanks{$^{2}$Amazon Inc.
        {\tt\small \{krsar, sgayaka, chkuo, senarnie\}@amazon.com}}%
\thanks{$^{3}$Google Inc.
        }
\thanks{Digital Object Identifier (DOI): see top of this page.}
}

\begin{document}

\maketitle

\begin{abstract}

    Object instance segmentation is a key challenge for indoor robots navigating cluttered environments with many small objects. Limitations in 3D sensing capabilities often make it difficult to detect every possible object. While deep learning approaches may be effective for this problem, manually annotating 3D data for supervised learning is time-consuming. In this work, we explore zero-shot instance segmentation (ZSIS) from RGB-D data to identify unseen objects in a semantic category-agnostic manner. We introduce a zero-shot split for Tabletop Objects Dataset (TOD-Z) to enable this study and present a method that uses annotated objects to learn the ``objectness'' of pixels and generalize to unseen object categories in cluttered indoor environments. Our method, SupeRGB-D, groups pixels into small patches based on geometric cues and learns to merge the patches in a deep agglomerative clustering fashion. SupeRGB-D outperforms existing baselines on unseen objects while achieving similar performance on seen objects. We further show competitive results on the real dataset OCID. With its lightweight design (0.4 MB memory requirement), our method is extremely suitable for mobile and robotic applications. Additional DINO features can increase the performance with a higher memory requirement. The dataset split and code is available at https://github.com/evinpinar/supergb-d.

\end{abstract}


\section{Introduction}
\label{sec:intro}

\IEEEPARstart{I}{nstance} segmentation is a long-standing and important problem with its applications from robotic perception to augmented reality \cite{maskrcnn, hafiz2020survey}. Especially for an indoor autonomous agent, detecting and identifying objects encountered in an environment is critical for visual navigation. However, the existing 3D sensing capabilities prevent detecting every possible object. Deep learning-based approaches are promising, but annotating 3D data is cumbersome and time-consuming. The problem is further exacerbated due to the variety of objects observed in a typical home environment. In this work, we address this issue by focusing on unseen object discovery in zero-shot from only a limited amount of annotated data.

Leveraging different sensor modalities such as RGB images and depth maps and carrying the task from 2D to 3D comes into further rescue when detecting object instances. Using only 2D cues (RGB) cannot capture the complexity of the 3D world, especially in high-textured areas \cite{DENG2017127}. Some methods use only depth, but they tend to result poorly when sensors fail (e.g. non-Lambertian surfaces, sharp edges) \cite{wang2018sgpn,liu2019point,depthbenchmark}. Few approaches use RGB-D, but they depend on large-scale annotated datasets or hand-crafted features \cite{wang2022tokenfusion}. Relying heavily on annotations is not scalable for an indoor agent, given the variations in the object categories encountered in different indoor environments \cite{simsar2021object}. Prior work aims to solve this task in sim2real \cite{xie2021unseen}, yet, employment of these methods could be difficult due to memory contraints.

\begin{figure}[t]
  \centering
   \includegraphics[width=0.99\linewidth]{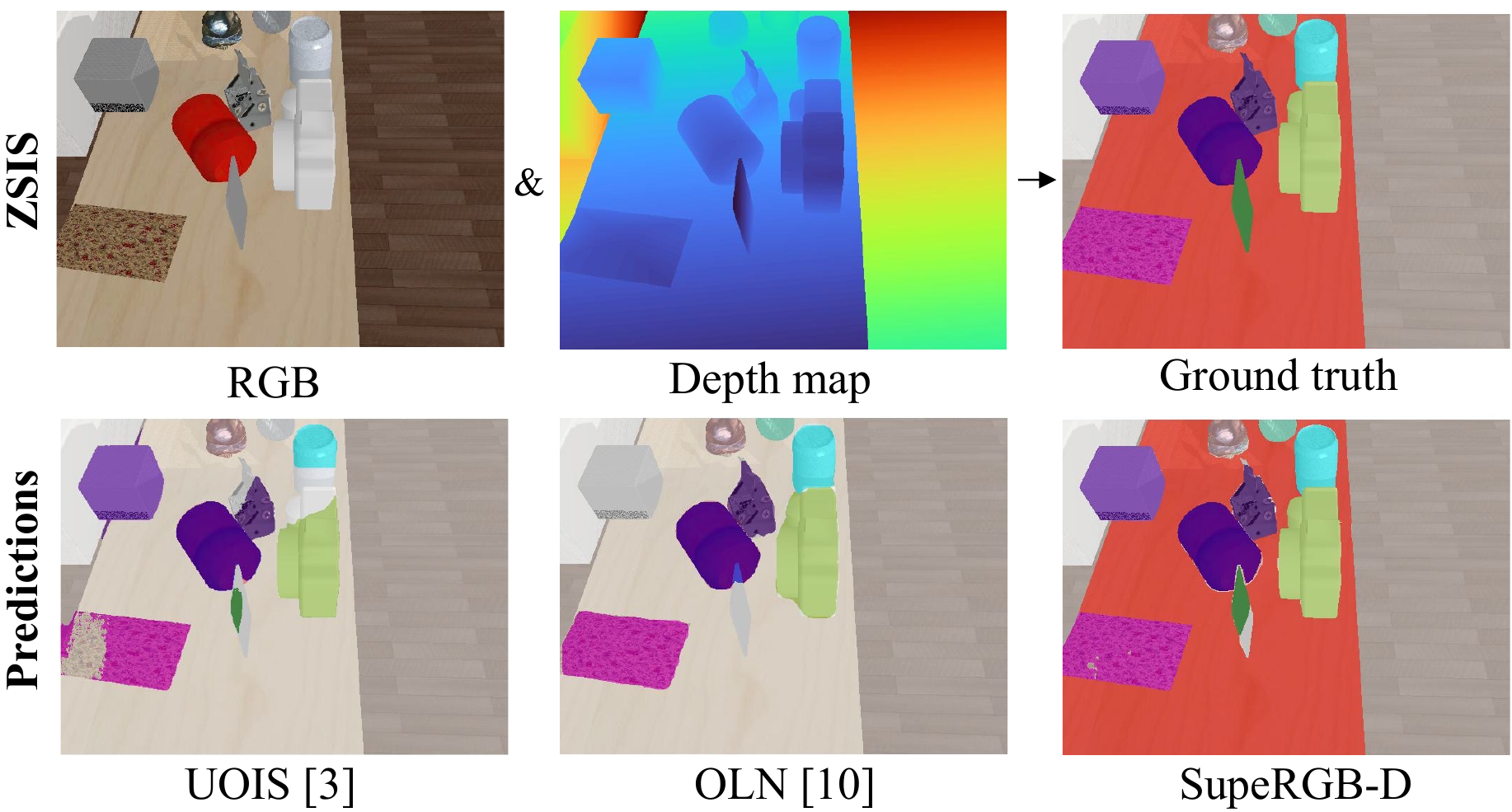}
   \caption{We study zero-shot instance segmentation from RGB and depth modalities, where a training split includes seen categories (\eg plate, mug), and the test split includes both seen and unseen categories (\eg camera, printer). We propose a dataset split and a method for ZSIS from RGB-D in cluttered environments. In this paradigm, compared to UOIS \cite{xie2021unseen} and OLN \cite{kim2021oln}, the proposed method SupeRGB-D segments unseen objects with accurate boundaries (camera, helmet), and can even identify things in the background (\eg table).}
   \label{fig:teaser}
\end{figure}

Towards this, we tackle the zero-shot instance segmentation (ZSIS) problem, where a training split can include objects from seen categories, whereas a test split with seen and unseen categories. We rely on RGB and depth modalities to solve this problem. As there are no available RGB-D datasets to study the problem under this multi-modal setup, we propose a novel dataset split, TOD-Z, where we divide the dataset into seen and unseen categories to enable a structured study into ZSIS. Through this dataset, we establish a benchmark and propose a suitable approach for the task. Our method, SupeRGB-D, first detects local regions from RGB images and depth maps by simultaneously applying super-pixel over-segmentation on the input modalities. Then, it builds instance masks in a bottom-up manner by learning to group these extracted small patches through visual and geometric cues in an agglomerative clustering fashion. We show that SupeRGB-D outperforms compared methods on unseen objects as illustrated in Fig. \ref{fig:teaser}, obtains strong boundaries due to super-pixels acquired edges, and is extremely suitable for mobile applications with its lightweight design. 

Our contributions can be summarised as follows: 
\begin{itemize} 
    \item We study zero-shot instance segmentation from RGB-D in an open-world scenario. Notably, it is one of the first works on generalization towards unseen categories in zero-shot paradigm as opposed to cross-data sim2real.
    \item To this end, we provide an in-depth analysis of the generalization performance under different setups and offer a zero-shot split over the original TOD, namely, the TOD-Z split. With this split, we establish a novel benchmark to enable studying ZSIS on RGB-D. 
    \item We propose a novel method, SupeRGB-D, that is explicitly tailored towards generalization for unseen objects under limited data. SupeRGB-D first over-segments the input into local patches, then learns to merge them to reflect objectness and infer instance mask. 
\end{itemize}

\section{Related Work}
\label{sec:related}

\myparagraph{Instance segmentation} is one of the fundamental tasks in computer vision \cite{hafiz2020survey}. Prior to the deep learning (DL) era, learning-free algorithms aimed to group the pixels according to a similarity in a bottom-up manner. These works used hand-crafted features, \eg, brightness, color, texture, and defined heuristics to learn the objectness. Some of these study segmentation in terms of graph partitioning and seek a global optimal solution (\cite{graphcut}, \cite{normalizedcut}, \cite{Felzenszwalb2004}). Recently, with the availability of large-scale datasets and advances in DL, the performance of instance segmentation improved \cite{pinheiro}.
The majority of methods first detect the object and then segment the detected region in a top-down manner. 
A nominal work is based on identifying probable object regions through Region Proposal Network (RPN) and then using predictors to classify and segment objects, \ie, Mask R-CNN \cite{maskrcnn}.
Albeit these methods perform remarkably, learning generalizable object recognition is data-hungry. 

\myparagraph{Low-shot segmentation} has gained importance to address the generalizability of instance segmentation to enable its application in real-world scenarios.
Zero-shot segmentation aims to densely segment input images along with the object labels for zero-shot objects, \ie, the objects that were never seen during training, through the usage of semantic cues such as WordNet embeddings \cite{XianCVPR2019b, bucher2019, zhao2022ncdss}. Few-shot simplifies the assumption by providing one or a few samples during training \cite{tian2022gfsseg, Ganea_2021_CVPR}. 
ZSIS aims to detect different instances of unseen categories. Zheng \etal studied ZSIS \cite{Zheng_2021_CVPR} through learning a semantic alignment on a Fast R-CNN instance predictor. Following, Kim \etal proposed Object Localization Network (OLN) \cite{kim2021oln} which turns off the object classifier of Mask R-CNN and learns a common objectness score. In an orthogonal direction, Wang \etal offered a ground-truth mask generation strategy to be used for detector supervision \cite{wang2022ggn}. We also study class-agnostic segmentation and aim to generalize beyond observed shapes. In contrast, we focus on cluttered indoor spaces for robotic applications and work on multi-modal RGB-D cues.

\myparagraph{Segmentation on RGB-D} modality became another possible direction of study after the availability of consumer level range sensors \cite{kinectfusion} and the relevant datasets \cite{nyu2012}. The segmentation can be applied on either on single-view RGB-D images \cite{ren_rgbd_1012}, or on full 3D point clouds which are reconstructed through multi-view RGB-D fusion algorithms \cite{liu2019point}. Initial works relied on hand-crafted features and heuristics such as normal, curvature and planarity \cite{ren_rgbd_1012,Deng_2015_ICCV,DENG2017127,gong}. Richtsfeld \etal generated patches with planar surface estimation and learned merging with SVM\cite{Richtsfeld}. 
Recent works use deep learning methods \cite{liu2019point, Hou_2019_CVPR, wang2022tokenfusion}. 
\textbf{\textit{Low-shot segmentation in 3D}} is a rather novel area of research \cite{zhao2021few,c3d,michele2021generative}. Some works looked into part discovery \cite{luo2020learning} or zero-shot semantic segmentation and classification for unseen 3D shapes \cite{c3d}. Meanwhile, in terms of robotics applications, previous methods looked into feature extraction recent trends perform \textbf{\textit{unseen object instance segmentation}} in a simulation to real data transfer setup \cite{danielczuk2019,back2020,xiang2020learning,laskey2021simnet,xie2021unseen, xie2021rice}. Xie \etal proposed UOIS-Net to learn objectness from RGB-D input, and provided with a large simulation dataset TOD. Back \etal predicts amodal instance segments \cite{back2022unseen}, while Durner \etal proposes a stereo dataset and a suitable method \cite{durner2021unknown}. As opposed to prior works, instead of cross-dataset sim2real paradigm, we study ZSIS.

\begin{figure}[t]
  \centering
   \includegraphics[width=0.75\linewidth]{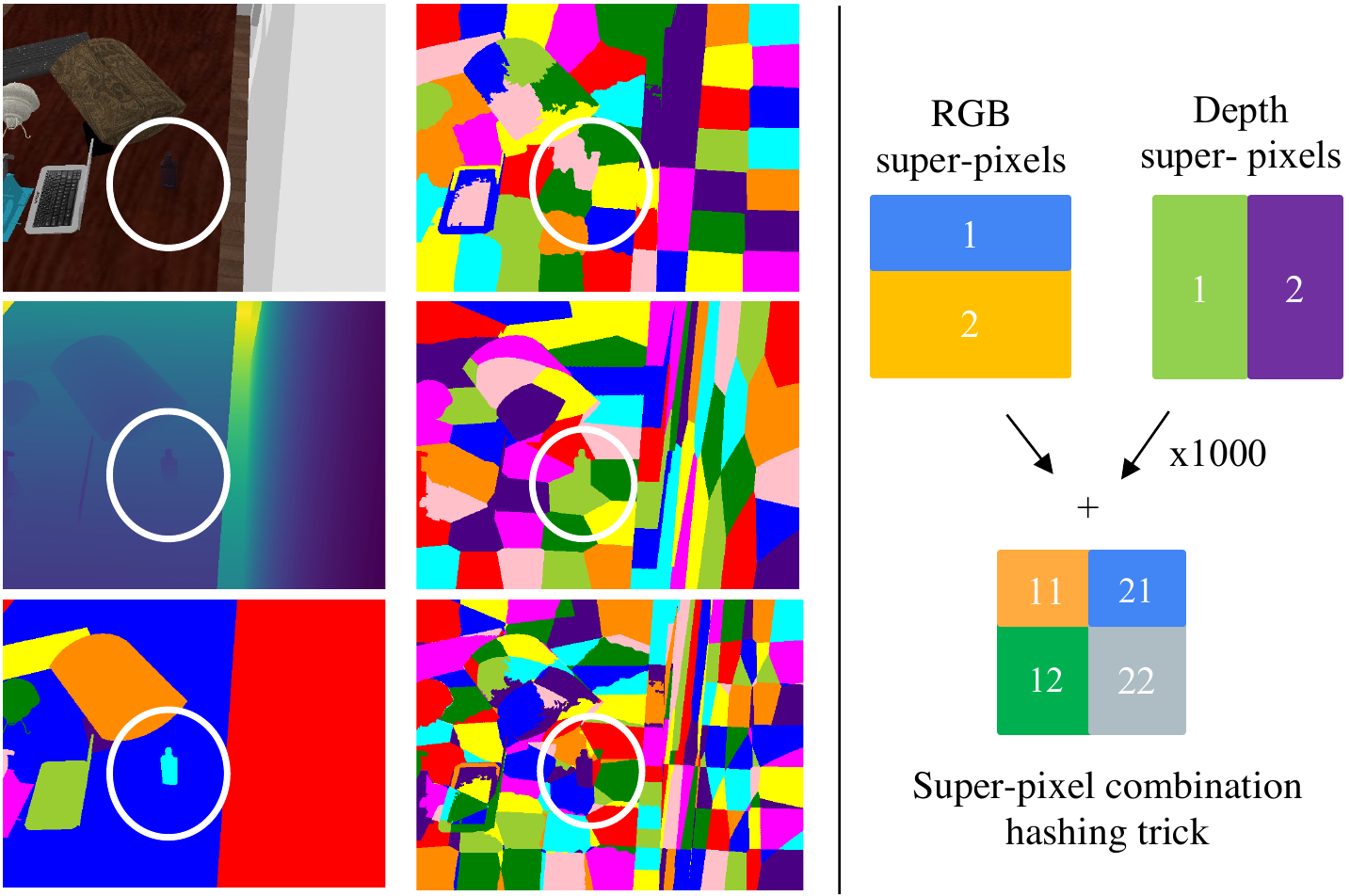}
   \caption{\textbf{RGB vs. depth modalities and super-pixels combination.} RGB and depth maps have complementary information. As depicted on the left, the bottle object was segmented out only by using both super-pixels. On the right, the super-pixels merging algorithm is illustrated.}
   \label{fig:slic_merging}
\end{figure}

\section{Method}
\label{sec:method}

\begin{figure*}[t!]
  \centering
   \includegraphics[width=0.89\linewidth]{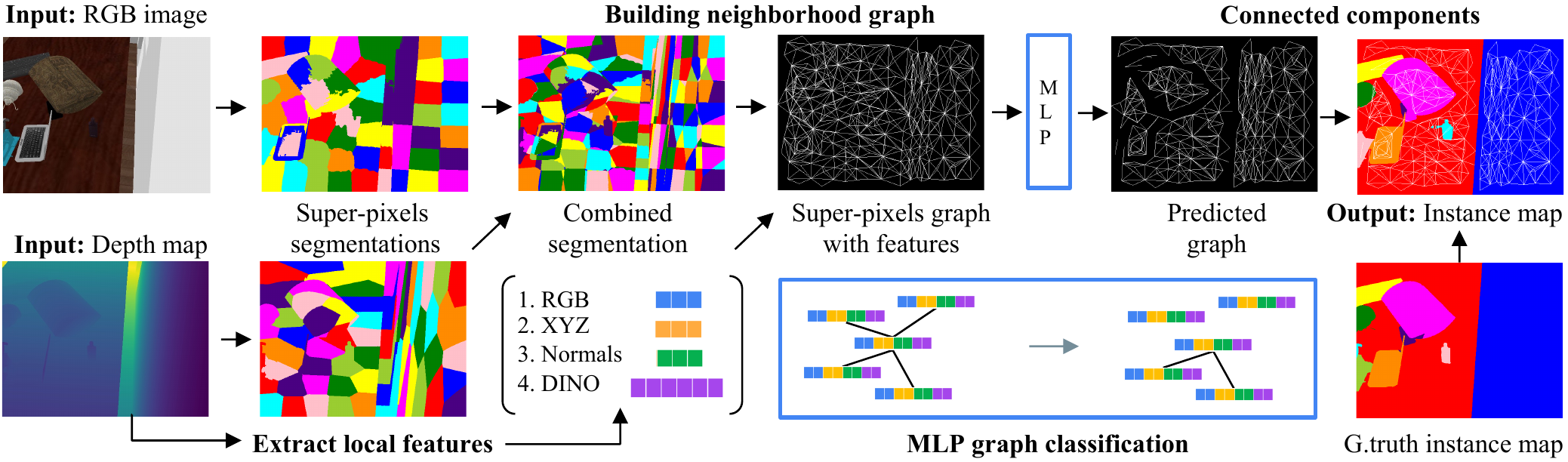}

   \caption{\textbf{Method illustration.} Our method SupeRGB-D first obtains local patches from input by applying super-pixels over-segmentation on RGB and depth maps separately and combining them to attain intersected areas, and the regions that lie in between. It then learns to merge the local patches to reflect the connectedness of objects by relying on extracted explicit (XYZ, normal and RGB) and implicit (DINO \cite{caron2021emerging}) features for patches and feeding each patch pair to a multi-layer perceptron. The final instance map is inferred by applying connected components on the predicted graph.}
   \label{fig:method}
\end{figure*}

In the following, we formalize the ZSIS problem and explain the proposed method. 

\myparagraph{Problem formulation.} 
We denote the set of seen classes as $\mathcal{S}$, a disjoint set of novel classes as $\mathcal{U}$ and the union of them as $\mathcal{Y} = \mathcal{S}\cup \mathcal{U}$. Let $\mathcal{T}=\{(x^{1}, x^{2}, y)| x\in \mathcal{X}_s, y\in\mathcal{Y}_s\}$, where $\mathcal{T}$ defines the training set, $x^{1}$ is an input image $I \in R^{3 \times H \times W}$ in the image space $\mathcal{X}_s$, $x^{2}$ is its input depth map $I \in R^{H \times W}$, $y$ is its class agnostic instance label mask with the same size. Training images include objects from seen object classes $\mathcal{S}$, whereas test images contain both seen and unseen classes from $\mathcal{U}$. Hence, the goal of ZSIS is to predict pixel-wise dense instance predictions among both seen and unseen classes. Note that, instead of semantic segmentation, we predict the class-agnostic instance segmentation map which enables open-world object discovery in real-world environments.

\myparagraph{Method overview.} 
We design a lightweight bottom-up strategy to learn instance segmentation from a small amount of data, without depending on object detection. We design a bottom-up strategy to infer instance segmentation which does not depend on object recognition and detection. Our algorithm first segments the input into small regions and then learns to merge these regions to infer a full instance segmentation map. Formally, we regard the segmentation task as a graph partitioning problem, where we initially generate an undirected graph $\mathcal{G} = (\mathcal{V}, \mathcal{E})$ on a super-pixels level to represent an image. A vertex $v_{i} \in \mathcal{V}$ represents one super-pixel region, and an edge $e = (v_{i}, v_{j}) \in \mathcal{E}$ connects two vertices $v_{i}$ and $v_{j}$. A segmentation solution $S$ is a partition of $\mathcal{V}$ into multiple connected components. To find the partitioning, as opposed to prior work (\cite{Felzenszwalb2004, graphcut}) that uses hand-crafted heuristics to find a threshold, our method learns the partitioning implicitly through a neural network. 

\subsection{Extracting local regions from RGB-D}

We aim towards a lightweight method for practical mobile and robot applications. Hence, we use the super-pixel over-segmentation instead of pixel-wise dense input to build our graph. Super-pixel methods can capture and summarize the most informative local regions from an image in an unsupervised manner. Furthermore, they provide strong boundaries reliably without requiring additional semantic input. As for the super-pixels algorithm, we select SLIC \cite{slic} because it is a simple yet robust approach, does not require any pretraining, has efficient runtime, and is accepted in the community for different downstream tasks \cite{spn2, simsar2021object}. 

\myparagraph{Combining superpixels from RGB and depth maps.} As illustrated in Fig. \ref{fig:slic_merging}, we apply SLIC separately on two input modalities, RGB and depth maps, resulting in two patch maps. Applying super-pixels on RGB and on depth maps has different indications. From RGB, it segments the regions according to color and brightness information. From the depth map, it understands the objectness of instances and the topology of surfaces through 3D cues. We aim to combine two maps to retrieve the intersecting areas between the patches and the regions lying in between. We use the following hashing trick: suppose each mask has N patches, where each patch is identified with a unique mask id, i.e. $P^{RGB}_{i}$ and $P^{depth}_{i}$, where $i \in R^{N}$.  For that, we multiply RGB superpixel ids with 1000 (shifting them three digits left) and add depth ids:
\begin{equation}
    P^{RGB}_{i} * 1000 + P^{depth}_{i}.
\end{equation}

\myparagraph{Explicit local features.} For each super-pixel region, we extract the local features that describe the properties of that local patch. As we aim to use the cues both from RGB and depth maps, we use the following features: (1) averaged R,G,B values within the patch, (2) X,Y,Z coordinates of the patch centroid, (3) surface normals of the centroid. For X,Y coordinates, we use image coordinates, whereas for Z, we use normalized depth value on that pixel $d_{i}$. We calculate the surface normal $n_{i}$ of the centroid pixel via vertical and horizontal gradients of the depth map as:
\begin{equation}
    n^{g}_{i} = [-\nabla_{x}(d_{i}), -\nabla_{y}(d_{i}), 1]^\top.
\end{equation}

\myparagraph{Implicit ViT features.} Explicit features already bring strong cues for patch merging, however, are limited in terms of global cues. To make the model globally aware, we rely on self-supervised Vision Transformer features from DINO\cite{caron2021emerging}. Notably, our goal is open-world generalizability, and we cannot rely on neural networks pretrained on large-scale datasets that already observe all categories (ImageNet ResNet \cite{resnet}). Self-supervised ViT, on the other hand, is trained under a self-distillation setup and complies with the zero-shot setup where no category annotations are needed. We extract attention maps from the ViT, average the features that lie within the regions and incorporate them as additional feature inputs in the super-pixel neighborhood graph. 

\noindent To sum up, a patch feature can be constructed by concatenating the explicit and implicit features in a vector:
\begin{equation}
\begin{split}
    F_{sp} = [R, G, B, X, Y, Z, n_{x}, n_{y}, n_{z}, \\ 
    DINO_{1}, ..., DINO_{M}]
\end{split}
\end{equation}
where $M$ denotes the number of ViT attention heads. This is a superset of features and we ablate the effect of each component in Experiments Section.

\subsection{Learning to merge}

In order to infer instance segments, our method learns to merge super-pixel regions through an agglomerative clustering fashion. After building a super-pixel neighborhood graph where neighboring patches are connected through an edge, we use a neural network to learn the merging of the patches by classifying the edge as a cut or no-cut criteria. For supervision, we generate a ground truth pair graph from the instance mask and use this mid-representation. 

\myparagraph{Model training.} We first pre-process the data to extract the super-pixels along with the explicit and implicit features. For learning to merge, we use shared Multilayer Perceptron  (MLP) layers, and as an input, we feed neighboring super-pixel patch pairs features $P_{i}, P_{j}$. The network is tasked to classify whether this patch pair is connected. We simply use binary cross-entropy loss to train the neural network:
\begin{equation*}
    L_{BCE}=-{(y\log(p) + (1 - y)\log(1 - p))}.
\end{equation*}
 
During training, we randomly swap half of the pairs to enable permutation invariance $P_{j}, P_{i}$. Furthermore, we realized that the positive/negative ratio of the data is naturally biased, as the boundary pixels on images occupy fewer pixels than the objects themselves. The training data indeed have $80\%$ positive samples, and this prevents the network to learn the negatives appropriately. Hence, we modify the sampling ratio to enable the learning of negative edges. We provide more details on this in the Experiments. 

\myparagraph{Inferring the instance segmentation map.} The model predicts binary edge labels representing which patch pairs are connected to each other. To infer the instance segmentation map, we need to identify disjoint connected sets from the patch neighborhood graph. As shown in Fig. \ref{fig:method}, we use connected-components algorithm for this purpose.

\section{Tabletop Objects Dataset for Zero-shot (TOD-Z)}

Zero-shot \textit{semantic} segmentation problem is well-studied with RGB image datasets \cite{XianCVPR2019b, spn2}, whereas there are only a limited number of works on zero-shot \textit{instance} segmentation \cite{wang2022ggn}. To systematically study ZSIS on RGB-D data, we believe that one should initially understand the effect of different modalities where sensor noise is minimized. Hence, to benchmark the RGB-D ZSIS framework, we select the synthetic Tabletop Object Dataset (TOD) \cite{xie2021unseen} and propose a zero-shot split upon this, namely, TOD-Z. TOD is composed of ShapeNet \cite{shapenet2015} objects cluttered on tabletop environments. There are 25 different categories of common household objects, such as \textit{bottle, helmet, microwave, pillow} with 40K scenes, containing 5 to 25 random objects randomly thrown on tables acquired from SUNCG environments \cite{song2016ssc}.  

\myparagraph{Data pre-processing.} The raw TOD data does not include the semantic segmentation masks. However, we need them to be able to identify the object classes and study the proposed task. We hence extract the Synset id of each CAD object from TOD metadata and extract semantic masks for each image. Furthermore, this data is created by setting up a scene and rendering multiple images for each scene from different viewpoints. Each scene contains one empty SUNCG home background image, as well as another image with an empty table placed in SUNCG home. Furthermore, some images contain only a single object occupying a very small fraction of an image (a couple of pixels). Therefore, we filter out these redundant and information-poor images. 

\begin{figure}[t]
  \centering
    \includegraphics[width=0.99\linewidth]{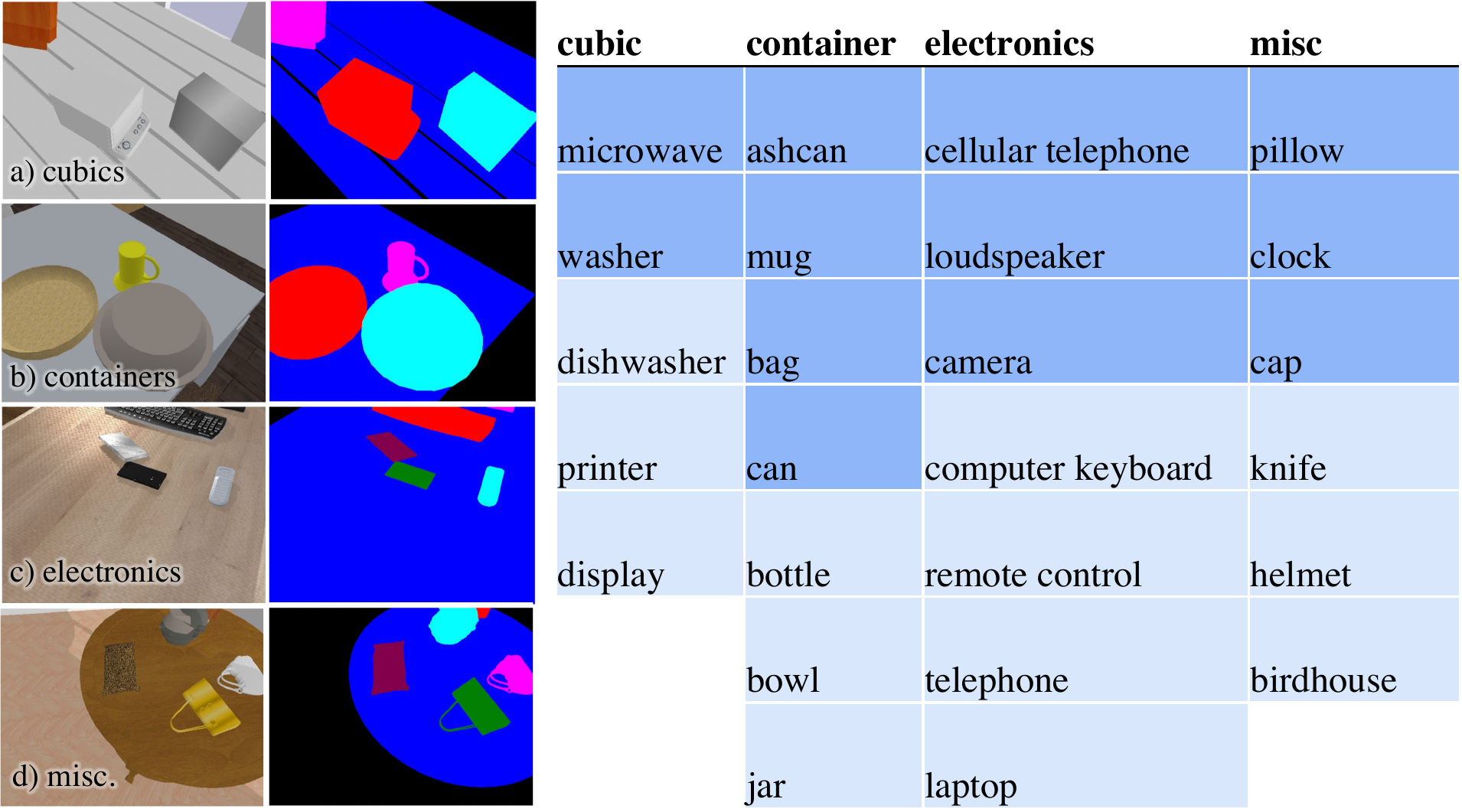}
   \caption{\textbf{Proposed TOD-Z Split}. We group the 25 object classes of TOD into 4 categories, and create a seen/unseen split by selecting equal numbers of classes from each category. The seen classes are highlighted with dark blue, whereas the unseen ones with light blue. }
   \label{fig:todz}
   \vspace{-0.3cm}
\end{figure}

\myparagraph{Selecting seen and unseen object categories.} For dividing the classes into seen (train) and unseen (test) classes, we follow a procedure similar to \cite{c3d,bansal2018zero}. The 25 object classes of TOD have a high variance in terms of both semantics and 3D structures. We initially group the classes in terms of semantics, by applying K-means clustering at the word2vec vectors as well as WordNet hierarchical distances. We further visually analyze images by extracting ResNet features for the objects appearing in images and checking similarities. Through this analysis, we group the object classes into four shape categories: containers (\eg, bowl, bottle, ashcan), cubics (\eg, washer, microwave), electronics (\eg, telephone, cellular), and miscellaneous (\eg, knife, birdhouse, pillow). We randomly select half of the classes from each group and assign them as seen, and the remaining as unseen. The final training split contains 12 classes, whereas the open test set contains 25 classes (12 seen \& 13 unseen), with 7118 train images and 881 test samples with equal object distributions. The TOD-Z classes and sample images can be seen in Fig. \ref{fig:todz}. We provide an in-depth analysis on of different classes and the number of samples in the Experiments.

\section{Experiments}
\label{sec:exp}

\begin{table}[t!]
  \centering
  \caption{\textbf{Comparing SupeRGB-D (our method) to baseline methods over harmonic mean, seen and unseen splits} in terms of object overlap precision, recall, and f-score, as well as boundaries. Higher values are better for all metrics. }

  \scalebox{0.9}{
  \begin{tabular}{c | c | c c c | c c c }
    \toprule
    \multirow{2}{*}{\textbf{Set}} & \multirow{2}{*}{\textbf{Method}} & \multicolumn{3}{c}{\textbf{Overlap}} & \multicolumn{3}{c}{\textbf{Boundary}}\\
     &  & \textcolor{orange}{P} & \textcolor{cyan}{R} & \textcolor{purple}{F} & \textcolor{orange}{P} & \textcolor{cyan}{R} & \textcolor{purple}{F} \\
    \midrule
    \multirow{3}{*}{\bfseries HM} & UOIS & 81.27 & 77.08 & 72.92 & 65.97 & \textbf{67.38} & 64.96 \\
    & OLN & 72.00 & \textbf{83.12} & 70.72 & 64.99 & 65.61 & 63.95 \\
    & SupeRGB-D & \textbf{89.67} & 74.44 & \textbf{76.48} & \textbf{69.17} & 67.99 & \textbf{66.75} \\
    \midrule
    \multirow{3}{*}{\bfseries Seen} & UOIS & 86.54 & 80.70 & \textbf{79.23} & \textbf{70.39} & \textbf{72.79} & \textbf{70.20} \\
    & OLN & 74.12 & \textbf{85.81} & 73.05 & 66.08 & 67.43 & 65.41 \\

    & SupeRGB-D & \textbf{90.35} & 73.69 & 76.31 & 66.37 & 64.16 & 63.71 \\
    \midrule
    \multirow{3}{*}{\bfseries Unseen} & UOIS & 76.60 & 73.78 & 67.53 & 62.08 & 62.73 & 60.45 \\
    & OLN & 69.99 & \textbf{80.60} & 68.53 & 63.94 & 63.88 & 62.55 \\
    & SupeRGB-D & \textbf{89.01} &  75.20 & \textbf{76.66} & \textbf{69.90} & \textbf{66.75} & \textbf{66.52} \\
    \bottomrule
  \end{tabular}
    }
  
  \label{tab:comparison}
\end{table}

\begin{table}[t!]
    \centering
    \caption{\textbf{Computational requirements} in terms of number of parameters in the neural network, model size in megabyte and runtime in seconds. SupeRGB-D has a lower number of parameters, fewer memory requirements, and a faster runtime than the compared baselines.}
    \scalebox{0.9}{
    \begin{tabular}{c|r|r|r}
    \toprule
        Method & \# Param. & Memory (MB) & Runtime [s]\\
        \midrule
        UOIS & 45.407.488 & 173.216 & 7.42 \\ 
        OLN & 43.744.526 & 167.935 & 4.50 \\
        SupeRGB-D & \textbf{103.682} & \textbf{0.400} & 1.44 \\ 
        \bottomrule
    \end{tabular}
    }
    \label{tab:model_size}
    \vspace{-0.4cm}
\end{table}

\myparagraph{Metrics.} We evaluate the \textbf{instance segmentation performance} with precision, recall and F-score metrics following the literature \cite{dave2019towards, xie2021unseen}. These metrics are calculated by first measuring the F-score between every ground truth and predicted instance mask pair, which are then matched with the Hungarian algorithm by the highest overlap F-score. The metrics between the matched objects are then calculated by:
\begin{equation}
P = \frac{\sum_i \left|s_i \cap g(s_i) \right|}{\sum_i \left|s_i\right|},
R = \frac{\sum_i \left|s_i \cap g(s_i) \right|}{\sum_j \left|g_j\right|},
F = \frac{2*P*R}{P+R} \nonumber
\end{equation}
where $s_i$ denotes the set of pixels belonging to predicted object $i$, $g(s_i)$ is the set of pixels of the matched ground truth object of $s_i$, and $g_j$ is the set of pixels for ground truth object $j$. We denote these metrics as Overlap P/R/F. Furthermore, we report the same numbers for the boundary pixels, where the boundaries are extracted by the edge detector and dilated. These metrics are calculated by considering the boundary pixels on the matched objects and they express how well the methods perform on fine-level details. Higher is the better for all of the metrics.

To measure the \textbf{zero-shot performance}, following the previous works \cite{XianCVPR2019b,c3d}, we report the instance segmentation metrics P/R/F averaged for seen classes, unseen classes and harmonic mean (HM) of seen\&unseen, which is computed as $HM = \frac{2*S*U}{S+U}$. The HM measures how well a model balances seen and unseen classes.

\myparagraph{Datasets.} We use TOD to examine the generalization gap between different classes as this synthetic dataset enables a structured study by precluding the effect of sensor noise. We propose TOD-Z split, where we demonstrate the usefulness of our method and compare it with others. To show SupeRGB-D performance on real images, we use the Object Clutter Indoor Dataset (OCID) \cite{ocid}, which is captured in real-world cluttered environments through a structured light depth sensor (ASUS-Pro) mounted on a robot.  

\begin{figure}[t!]
  \centering
   \includegraphics[width=\linewidth]{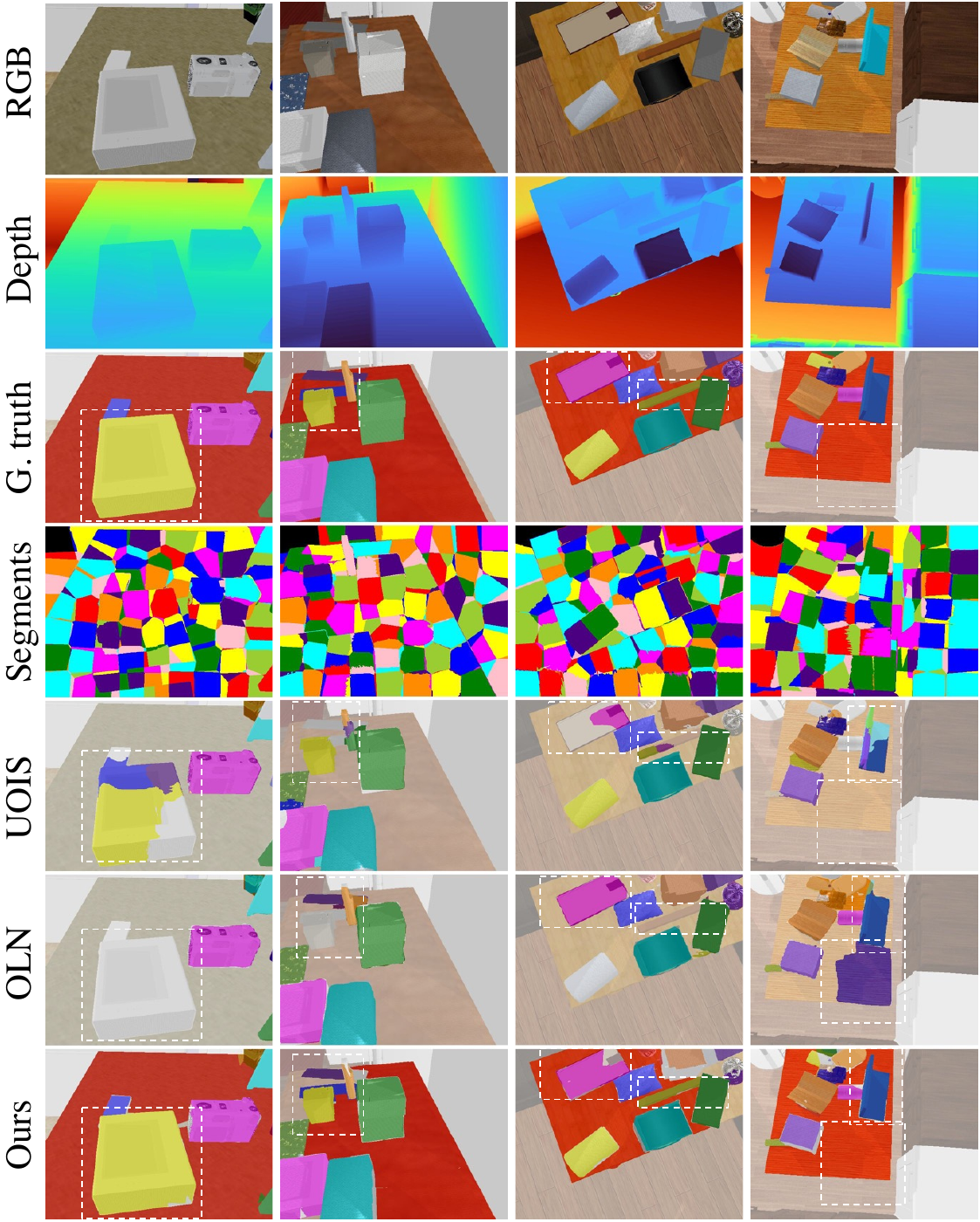}
   \caption{\textbf{Qualitative results.} Samples in cluttered environments with unseen object categories varying from the dishwasher, bottle, keyboard, display, and laptop (left to right). SupeRGB-D can segment these instances, whereas the compared methods fail to detect the objects as they tend to overfit into seen categories. Further, we can identify distinct objects in cluttered scenes captured from different perspectives such as floor parallel view (left two) or top view (right two), and even the table and objects beyond the background (right bottom). }
   \label{fig:qual}
   \vspace{-0.3cm}
\end{figure}

\myparagraph{Implementation details.} We implement the merging network with a 3-layer MLP of latent feature sizes $(256, 1024, 256)$, each followed by ReLU, and dropout layers. We use Adam optimizer with a step size learning rate scheduler on an initial learning rate $1e^{-3}$ and a reduction rate of 0.5. We train our network for 10 epochs and select the checkpoint with the highest validation F-score. For over-segmentation, we use the SLIC algorithm, where we find the ideal patch size of 128 after ablation studies.

\myparagraph{Baseline methods.} We select two strong methods that aim toward detecting and segmenting unseen objects. The first one is the aforementioned UOIS-Net method, which was tailored for RGB-D segmentation within sim2real \cite{xie2021unseen}. Their method consists of two steps, first using the depth map to detect the object instances, followed by improving the object boundaries through RGB image. We train it with the proposed TOD-Z split to measure the ZSIS performance. Our second baseline is Object Localization Network (OLN) \cite{kim2021oln}, which was built upon Mask R-CNN \cite{maskrcnn}, but aims to detect in an open-world setup by replacing the RPN object classification head with a generic object understanding head. After modifying this network for taking RGB-D input, we train it with TOD-Z. We follow the public code and training details provided by the authors.

\begin{table}[t!]
  \centering
  \caption{\textbf{Data sampling studies.} Increasing the negative ratio helps the model learn objects better, increasing all metrics. The best positive/negative ratio is $25/75$. }
  \scalebox{0.84}{
  \begin{tabular}{c | c | c c c}
    \toprule
    \multirow{2}{*}{\textbf{Set}} & \multirow{2}{*}{\textbf{p/n ratio}} & \multicolumn{3}{c}{\textbf{Overlap}} \\
     &  & \textcolor{orange}{P} & \textcolor{cyan}{R} & \textcolor{purple}{F} \\
    \midrule
    \multirow{4}{*}{\bfseries HM} & Random & 55.89 & 66.32 & 51.25\\
    & $50/50$ & 65.15 & 69.07 & 59.55 \\
    & $25/75$ & 86.91 & \textbf{78.71} & \textbf{78.43} \\
    & $10/90$ & \textbf{93.37} & 66.76 & 73.64 \\
    \midrule
    \multirow{4}{*}{\bfseries Seen} & Random & 55.29 & 67.23 & 50.94 \\
    & $50/50$ & 64.49 & 69.83 & 59.24 \\
    & $25/75$ & 87.66 & \textbf{79.16} & \textbf{79.11} \\
    & $10/90$ & \textbf{94.24} & 66.23 & 75.53 \\
    \midrule
    \multirow{4}{*}{\bfseries Unseen} & Random & 56.45 & 65.43 & 51.56 \\
    & $50/50$ & 65.82 & 68.33 & 59.86 \\
    & $25/75$ & 86.18 & \textbf{78.26} & \textbf{77.76} \\
    & $10/90$ &  \textbf{92.53} & 67.29 & 73.74 \\
    \bottomrule
  \end{tabular} 
    }
  \label{tab:pnsampling}
  \vspace{-0.5cm}
\end{table}

\subsection{Comparison to baselines}

We \textbf{quantitatively} compare our method without DINO features against the baselines in terms of instance segmentation performance in Tab. \ref{tab:comparison} and in terms of computational requirements in Tab. \ref{tab:model_size}. Compared to baselines, SupeRGB-D results with the best overlap \fscore and \precision on harmonic mean and unseen splits and the best boundary scores on all splits. On seen objects, SupeRGB-D has the highest \precision, whereas UOIS has the highest \fscore, showing that this method is strong on seen objects and is better than OLN probably because their model design considers 3D cues explicitly. On unseen objects, SupeRGB-D achieves nearly $10\%$ improvement on \fscore and boundary scores over the compared baselines. This shows that the baseline methods overfit into seen object categories and are rather limited in generalization over unseen types of shapes.  
In terms of \textbf{model size}, SupeRGB-D has fewer neural network parameters and has lower memory requirements than the compared methods. We furthermore compare in terms of runtime efficiency in seconds, where we measure the runtime of our method, including the data preprocessing (superpixels extraction) and postprocessing (segmentation inference via connected components) on an i7-8700 3.20GHz 12-core CPU. SupeRGB-D has the fastest runtime with 1.44 seconds, followed by 4.50 OLN, which is three times longer than SupeRGB-D, and 7.42 UOIS, with the longest probably due to data preprocessing. The lightweight design of SupeRGB-D enables running it on any robot or a mobile device with low-cost hardware. Our method's performance can be improved by additional ViT features with an increased memory requirement as we show in ablations.

We further provide a \textbf{qualitative} comparison to baselines in Fig. \ref{fig:qual}. We provide samples with unseen objects, dishwasher, bottle, keyboard, display, and laptop, captured from different viewpoints (bottom and top). It can be observed that compared methods perform poorly on these objects, either cannot detect them or hallucinate a couple of different objects on the overlapping pixels. Comparably, SupeRGB-D can segment and identify the objects as it relies on local cues and does not overfit into categories. Interestingly, it can even detect the objects beyond the categories, that lie in the background such as the table.

\begin{table}[t!]
  \centering
  \caption{\textbf{Features ablation.} We evaluate SupeRGB-D by ablating different feature inputs.}
  \scalebox{0.899}{
  \begin{tabular}{ c c c c | c c c | c c c }%
    \toprule
    \multicolumn{4}{c}{\textbf{Input}} & \multicolumn{3}{c}{\textbf{Overlap}} & \multicolumn{3}{c}{\textbf{Boundary}}\\
    RGB & D & N & DINO
       & \textcolor{orange}{P} & \textcolor{cyan}{R} & \textcolor{purple}{F} & \textcolor{orange}{P} & \textcolor{cyan}{R} & \textcolor{purple}{F} \\
    \midrule
    \cmark & & & & 76.51 & 75.81 & 69.51 & 60.42 & 68.95 & 61.91 \\
     & \cmark & &  & 53.49 & 62.95 & 48.26 & 35.20 & 37.24 & 34.24 \\ 
     &  $\mkern3mu$ \cmark$^\ast$ & & & 61.62 & 60.26 & 52.62 & 36.79 & 36.50 & 34.81 \\ 
     & & & \cmark & \textbf{94.46} &  55.67 & 65.47 &  58.04 &  46.56 &  50.26 \\
     & $\mkern3mu$ \cmark$^\ast$ & & \cmark & 88.26 & 63.37 & 67.64 & 59.22 & 51.68 & 53.25\\
     \cmark & \cmark & &  & 87.62 & 75.33 & 76.33 & 68.09 & 65.42 & 65.08\\
\cmark & \cmark & \cmark & & \textbf{89.67} & 74.44 & 76.48 & 69.17 & 67.99 & 66.75\\
    \cmark & \cmark & \cmark &  \cmark &  86.91 & \textbf{78.71} & \textbf{78.43} & \textbf{72.05} & \textbf{71.23} & \textbf{69.69} \\
    \bottomrule
  \end{tabular}%
    }
  \label{tab:feat_abl}
  \vspace{-0.3em}
\end{table}

\begin{figure}[t!]
  \centering
   \includegraphics[width=0.75\linewidth]{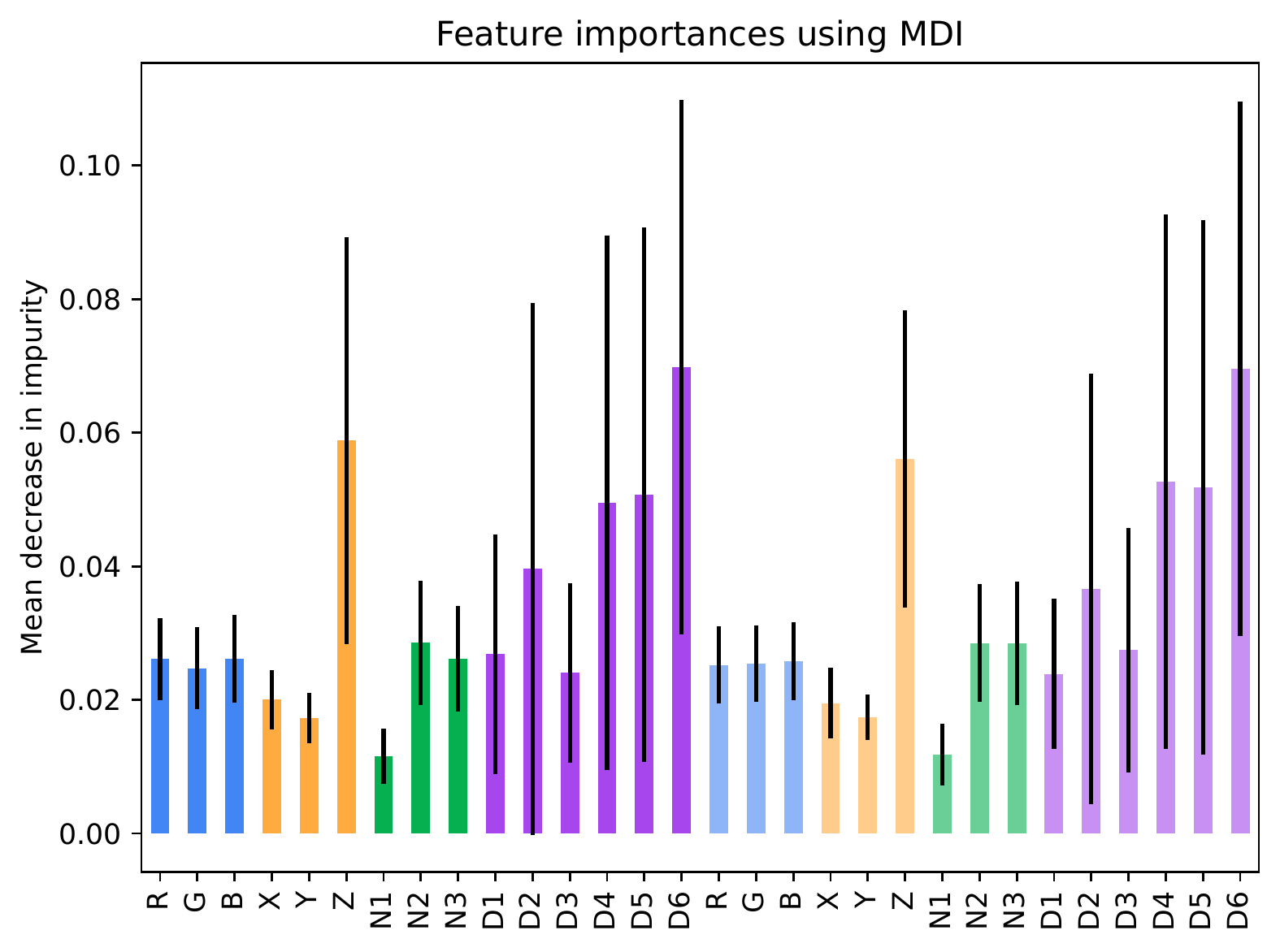}
   \caption{\textbf{Feature importances for an input patch pair.} Depth values (Z) and DINO6 (ViT 6th attention head) features have the highest importance on a Random Forest Classifier.}
   \label{fig:rfi}
   \vspace{-0.5cm}
\end{figure}

\vspace{-0.3em}
\subsection{Ablation studies}

\myparagraph{Positive/negative sampling.} SupeRGB-D requires balancing the positive and negative edge ratios to encounter the data imbalance, \ie, boundary vs. in-object pixels. We compare the default sampling, where the input reflects the dataset ratio of $80\%$ positives and modified ratios of $50/50$, $25/75$, and $10/90$ p/n in Tab. \ref{tab:pnsampling}. There is a clear correlation between the negative ratio and performance. First, having a $50/50$ ratio keeps \precision similar, but it increases \recall significantly by around $10\%$. Further, increasing the negative ratio to $25/75$ boosts both \precision and \recall. More increase, however, results in a drop in \recall and \fscore, probably due to over-weighting the edge pixels.

\myparagraph{Effect of features.} SupeRGB-D relies on features from different modalities, namely, RGB, XYZ, normals, and DINO features. We compare the effect of each in Tab. \ref{tab:feat_abl} on the harmonic mean. We observe that (1) using only RGB features enables class-agnostic instance segmentation, whereas (2) using only depth does not give strong enough cues for the task. (3) Combining RGB and depth features improve upon the RGB on all overlap and boundary metrics. (4) The addition of normals does not significantly affect overlap scores, but it helps with boundaries. (5) The full model with DINO features boosts the performance further.

\myparagraph{Feature importance.} To understand the effect of features out of a black box model, we fit a Random Forest classifier and analyze the importance of each input feature on the output in Fig. \ref{fig:rfi}. For this study, we randomly select 512 patches and fit a classifier. The feature importance is calculated as the decrease in node impurity weighted by the probability of reaching that node \cite{rf2001}. Higher values show higher importance in the model decision. The Z and DINO 6 features have the highest effect on the classification outcome. 

\myparagraph{Superpixel size.} We compare superpixel sizes of 32, 64, and 128 in Tab. \ref{tab:sp}. Increasing the patch size from 32 to 64 has significant gains on all overlap and boundary metrics (\eg $12\%$ on overlap, $13\%$ on boundary \fscore). This is expected, as a smaller number of patches don't cover objects well and result in missing objects. Doubling 64 to 128 only helps in boundary \fscore, as finer patches cover boundaries better. 

\begin{table}[t!]
  \centering
  \caption{\textbf{Quantitatives on OCID.} Sim2real (S2R) and ZSIS setup comparison on YCB10.}
  \scalebox{0.80}{
  \begin{tabular}{ c c c | c c c | c c c }%
    \toprule
    \multicolumn{3}{c}{\textbf{Method}} & \multicolumn{3}{c}{\textbf{Overlap}} & \multicolumn{3}{c}{\textbf{Boundary}}\\
     & S2R & ZSIS & \textcolor{orange}{P} & \textcolor{cyan}{R} & \textcolor{purple}{F} & \textcolor{orange}{P} & \textcolor{cyan}{R} & \textcolor{purple}{F} \\
    \midrule
    UOIS & \cmark & & 51.73 & 34.26 & 30.86 & 25.05 & 27.83 & 23.22 \\
    OLN & \cmark & & 47.79	& 15.90	& 17.46	& 34.81	& 14.94	& 17.66 \\
    Ours & \cmark & & 59.02 & 47.09 & 38.44 & 36.12 & 32.50 & 30.42 \\
    \midrule
    OLN & & \cmark& 58.31 & 46.48 &  38.21 &  46.44 & 50.82 & 43.12 \\ 
    Ours & &  \cmark& 55.46 & 60.22	& 51.23	& 41.52 & 55.92	& 45.65\\ 
    \bottomrule
  \end{tabular}%
    }
  \label{tab:ocid_quant}
\end{table}

\begin{figure}[t!]
  \centering
   \includegraphics[width=0.8\linewidth]{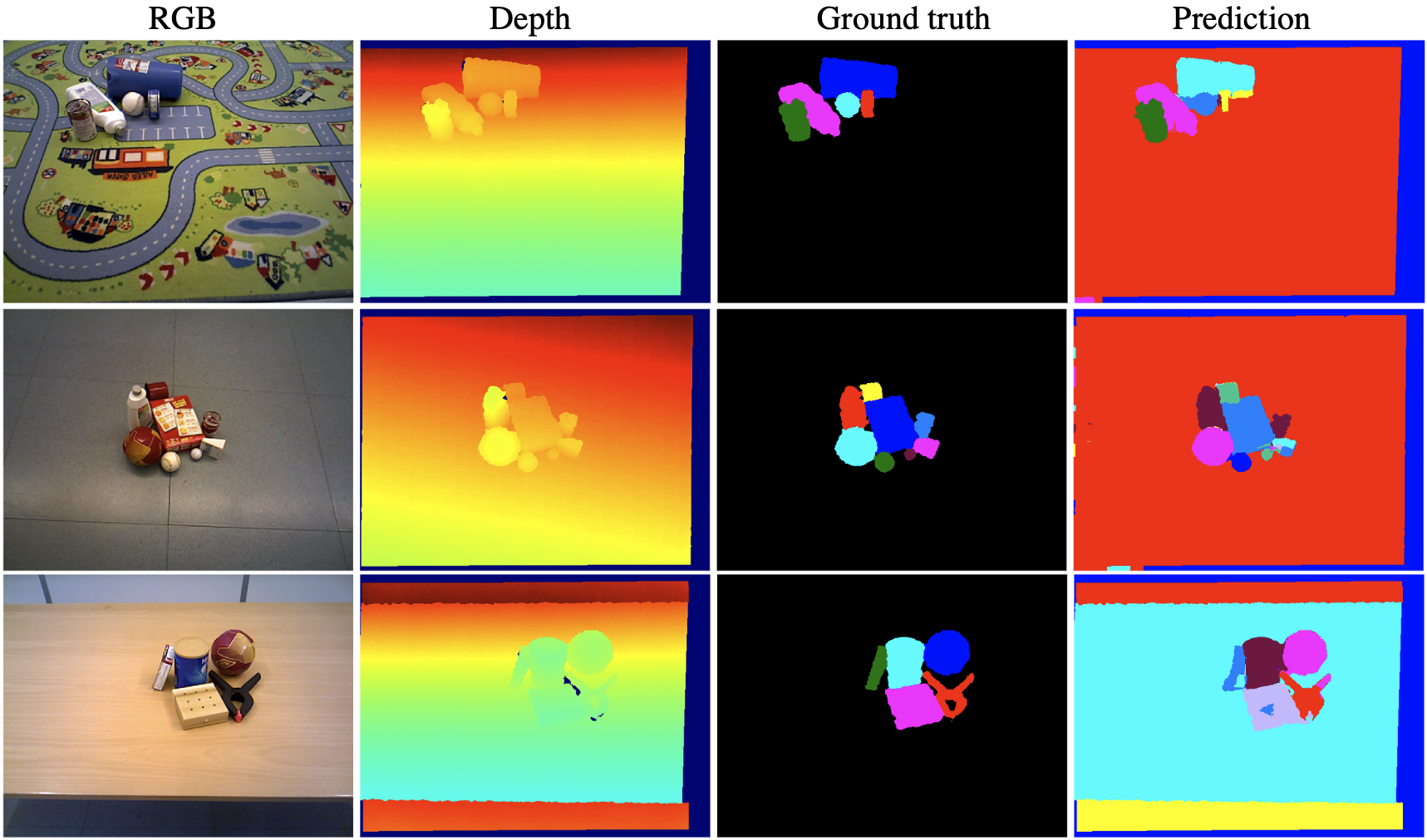}
   \caption{Results on real-world RGB-D dataset OCID YCB10 split, samples from floor and tabletop environments. }
   \label{fig:ocid_results}
   \vspace{-1.0em}
\end{figure}

\subsection{Results on real-world images}
As of today, there are no zero-shot splits available on real-world RGB-D datasets. Yet, to test SupeRGB-D capability under real sensor depth maps, we provide results on the OCID dataset in Fig. \ref{fig:ocid_results} and Tab. \ref{tab:ocid_quant}. This dataset consists of various types of objects in different tabletop and floor environments of different materials. For complementary reasons, we first test on sim2real setup with the TOD-Z trained models. Then for the ZSIS problem, we train SupeRGB-D and OLN methods with ARID objects and test on YCB10 objects. SupeRGB-D performs well on high-texture images, as well as noisy depth maps, and segments things in the background. 

\vspace{-0.3em}
\subsection{Generalization studies on TOD}

\begin{table}[t!]
  \centering
  \caption{\textbf{Number of super-pixels.} We compare the effect of patch sizes in terms of the number of super-pixels in the over-segmentation mask and provide results on the harmonic mean. Increasing the number of super-pixels results in better instance segmentation.}
  \scalebox{0.89}{
  \begin{tabular}{ c | c c c | c c c }%
    \toprule
    \multirow{2}{*}{\textbf{\# patches}} & \multicolumn{3}{c}{\textbf{Overlap}} & \multicolumn{3}{c}{\textbf{Boundary}}\\
       & \textcolor{orange}{P} & \textcolor{cyan}{R} & \textcolor{purple}{F} & \textcolor{orange}{P} & \textcolor{cyan}{R} & \textcolor{purple}{F} \\
    \midrule
    
    32 & 71.70 & 74.09 & 66.03 & 59.65 & 63.20 & 56.61 \\
    64 & 86.91 & 78.71 & 78.43 & 72.05 & 71.23 & 69.69 \\
    128 & \textbf{88.80} & 78.79 & 78.46 & \textbf{74.02} & 71.84 & 70.89 \\
    256 & 87.89 & \textbf{80.94} & \textbf{80.31} & 73.30 & \textbf{73.40} & \textbf{71.67} \\
    
    \bottomrule
  \end{tabular}%
    }
  \label{tab:sp}
\end{table}

\begin{table}[t!]
    \caption{\textbf{Dataset generalization gap studies.}}
    \begin{subtable}{.5\linewidth}
      \centering
        \resizebox{0.99\linewidth}{!}{
        \begin{tabular}{c | c | c c c }%
    \toprule
    \textbf{\# obj} & \textbf{set} & \textcolor{orange}{P} & \textcolor{cyan}{R} & \textcolor{purple}{F} \\
    \midrule
    \multirow{2}{*}{15} & seen & 83.04 & 87.29 & 84.09\\
    & unseen &  75.50 & 83.69 & 77.50 \\
    \midrule
    \multirow{2}{*}{20} & seen & 88.89 & 89.30 & 88.69\\
    & unseen &  86.37 & 88.55 & 85.91\\
    \midrule
    \multirow{2}{*}{23} & seen & 93.45 & 91.29 & 91.94 \\
    & unseen & 84.37 & 84.36 & 82.27 \\
    \bottomrule
  \end{tabular}
  }
  \caption{}
    \end{subtable}%
    \begin{subtable}{.5\linewidth}
      \centering
        \resizebox{0.99\linewidth}{!}{
        \begin{tabular}{c | c | c c c }%
    \toprule
    \textbf{\#data} & \textbf{\#obj} & \textcolor{orange}{P} & \textcolor{cyan}{R} & \textcolor{purple}{F}  \\
    \midrule
    \multirow{2}{*}{2070} & 5 & 67.91 & 52.37 & 69.37  \\
     & 25 & 80.83 & 67.93 & 71.61\\
    \midrule
    \multirow{2}{*}{6375} & 10 & 72.19 & 64.62 & 76.03 \\
     & 25 & 84.96 & 76.72 & 79.26 \\
    \midrule
    \multirow{2}{*}{17265} & 15 & 81.84 & 75.42 & 85.81 \\
     & 25 &  86.94 & 83.09 & 90.31\\
    \bottomrule
      \end{tabular}
      }
      \caption{}
    \end{subtable}
    \begin{subtable}{.5\linewidth}
      \centering
        \resizebox{0.99\linewidth}{!}{
        \begin{tabular}{c | c | c c c }%
    \toprule
    \textbf{train} & \textbf{Set} & \textcolor{orange}{P} & \textcolor{cyan}{R} & \textcolor{purple}{F}\\
    \midrule
    \multirow{2}{*}{15-var1} & seen &  83.04 & 87.29 & 84.09\\
    & unseen &  75.50 & 83.69 & 77.50 \\
    \midrule
    \multirow{2}{*}{15-var2} & seen & 82.44 & 84.96 & 82.32\\
    & unseen &  78.4 & 82.62 & 79.00 \\
    \midrule
    \multirow{2}{*}{15-var3} & seen & 86.15 & 84.43 & 85.84\\
    & unseen & 80.14 & 84.97 & 80.86 \\
    \bottomrule
  \end{tabular}
  }
  \caption{}
    \end{subtable}%
    \begin{subtable}{.5\linewidth}
      \centering
        \resizebox{0.99\linewidth}{!}{
        \begin{tabular}{c | c | c c c }%
    \toprule
    \textbf{train} & \textbf{Set} & \textcolor{orange}{P} & \textcolor{cyan}{R} & \textcolor{purple}{F} \\
    \midrule
    \multirow{2}{*}{cubics} & seen & 90.50 & 77.84 & 81.31 \\
    & unseen & 71.94 & 54.80 & 58.93 \\
    \midrule
    \multirow{2}{*}{cont.} & seen &  91.50 & 80.41 & 83.55 \\
    & unseen & 68.42 & 52.24 & 55.88 \\
    \midrule
    \multirow{2}{*}{electr.} & seen &  79.91 & 66.20 & 68.43\\
    & unseen &  71.89 & 49.95 & 55.51 \\
    \midrule
    \multirow{2}{*}{misc.} & seen & 82.26 & 78.35 & 77.35 \\
    & unseen &  79.53 & 69.07 & 71.29 \\
    \bottomrule
  \end{tabular}
      }
      \caption{}
    \end{subtable} 
    \label{tbl:data_studies}
    \vspace{-1cm}
\end{table}

Before proposing TOD-Z split, we conducted a set of experiments to study the generalization gap between different seen and unseen classes on the original TOD. We split TOD into seen train, seen test, and closed unseen test, and run experiments on UOIS method. The closed unseen test includes images of only unseen classes as opposed to open-world split that can have both seen+unseen objects. We report Overlap scores. 

\myparagraph{1. Seen/unseen gap for different number of seen classes. } We prepare three training splits by selecting images of 15/20/23 seen objects and the remaining (10/5/2) unseen. As shown in Tab. \ref{tbl:data_studies} (a), there is a generalization gap between seen and unseen data. When there are more training classes, the performance on both seen and unseen splits are higher.

\myparagraph{2. Number of train samples versus class variance.} We prepare three training splits with a varying number of data samples (2070/6375/17265) and either have a set of seen classes(5/10/15) or include all 25 classes. We report the values on the public TOD test split, which includes objects from 25 classes. In Tab. \ref{tbl:data_studies} (b), increasing class variance helps on generalizability.

\myparagraph{3. Generalization gap for different class types.} We prepare three training datasets of 15 seen classes, where each dataset has the same number of samples but different 15 categories. In Tab. \ref{tbl:data_studies} (c), we report the generalization gaps between closed seen and unseen test splits. We see that types of classes matter, meaning that an ideal training dataset should contain a variety of objects to generalize better. 

\myparagraph{4. Proposed TOD-Z groups on generalizability.} As explained in Section 4, the original TOD can be grouped into four groups after the similarity analysis. We prepare four training sets for each group and measure the generalization gap. In Tab. \ref{tbl:data_studies} (d), electronics and miscellaneous objects are more difficult to learn, as the seen performances are generally lower than the other categories. Yet, they perform better on unseen objects, indicating that training with these samples helps with generalizability.

\section{Conclusion and Future Works}
\label{sec:conc}

In this work, we study the problem of ZSIS from RGB-D images for cluttered indoor environments in a class-agnostic manner. Towards that, we analyze the generalization gap on an RGB-D dataset TOD and nominate a novel zero-shot RGB-D split, TOD-Z. We propose SupeRGB-D tailored for ZSIS, which first extracts small patches from RGB-D images and then learns to merge them through local visual and 3D cues. Our approach outperforms the baselines on unseen instances and is very lightweight with a 0.4 MB memory requirement that can be used on mobile robots and embedded devices. Future work includes incorporating uncertainty estimates to reflect sparsity and noise characteristics of real depth maps when fusing two modalities to improve the accuracy of resulting segmentation.

\addtolength{\textheight}{-12cm}   %

\bibliography{egbib}

\end{document}